\pdfoutput=1

\documentclass[11pt]{article}

\usepackage[final]{acl}

\usepackage{mathtools} 
\usepackage{makecell}
\usepackage{times}
\usepackage{latexsym}
\usepackage{amsmath}   
\usepackage{amsfonts}  
\usepackage{booktabs}  
\usepackage{listings}
\usepackage{xcolor}
\usepackage{mhchem} 
\DeclareUnicodeCharacter{2713}{\checkmark}

\usepackage[most]{tcolorbox}  
\usepackage{algorithm2e}
\usepackage{amssymb}     
\usepackage{verbatim}    
\usepackage{xcolor}      
\usepackage{hyperref}    

\usepackage{subfigure}
\usepackage{booktabs} 
\usepackage{colortbl}
\usepackage{multirow}
\usepackage{bbm}
\usepackage{enumitem}
\usepackage{makecell}

\usepackage[T1]{fontenc}

\usepackage[utf8]{inputenc}
\usepackage{amsmath}
\usepackage{amsthm}
\usepackage{enumitem}
\usepackage{multirow}
\usepackage{subcaption}
\usepackage{subfloat}
\usepackage{booktabs}

\usepackage{microtype}

\usepackage{inconsolata}

\usepackage{graphicx}

\usepackage{enumitem}
\usepackage{listings}  
\usepackage{pifont}
\usepackage[textsize=tiny]{todonotes}

\definecolor{darkgrey}{rgb}{0.53,0.53,0.53}
\definecolor{mygrey}{rgb}{0.9,0.9,0.9}
\definecolor{purple}{RGB}{230, 227, 254}
\definecolor{lightgreen}{RGB}{238, 252, 241}
\definecolor{lightred}{RGB}{231, 187, 187}
\definecolor{darkred}{RGB}{198, 129, 129}

\definecolor{tabhighlight}{HTML}{e5e5e5}
\definecolor{someorange}{rgb}{0.773,0.353,0.067}
\definecolor{someblue}{rgb}{0.27, 0.35, 0.760}

\title{Reinforcement Learning for Tool-Integrated Interleaved Thinking towards Cross-Domain Generalization}

\author{
Zhengyu Chen$^{1}$\footnotemark[1], 
Jinluan Yang$^{2}$\footnotemark[1],
Teng Xiao$^{3}$,
Ruochen Zhou$^{4}$, 
Luan Zhang$^{1}$ \\
Xiangyu Xi$^{1}$, 
Xiaowei Shi$^{1}$, 
Wei Wang$^{1}$, 
Jinggang Wang$^{1}$ \\
$^1$ Meituan \\
$^2$ Zhejiang University \\
$^3$ Allen Institute for Artificial Intelligence \\
$^4$ City University of Hong Kong \\
\texttt{chencsmat@gmail.com, yangjinluan@zju.edu.cn}
}

\begin{document}
\maketitle

\footnotetext[1]{Equal Contribution.}

\begin{abstract}
Recent advances in large language models (LLMs) have demonstrated remarkable capabilities in reasoning and tool utilization. However, the generalization of tool-augmented reinforcement learning (RL) across diverse domains remains a significant challenge. Standard paradigms often treat tool usage as a linear or isolated event, which becomes brittle when transferring skills from restricted domains (e.g., mathematics) to open-ended tasks. In this work, we investigate the cross-domain generalization of an LLM agent trained exclusively on mathematical problem-solving. To facilitate robust skill transfer, we propose a {\textbf{R}einforcement Learning for \textbf{I}nterleaved \textbf{T}ool \textbf{E}xecution (RITE)}. Unlike traditional methods, RITE enforces a continuous ``Plan-Action-Reflection'' cycle, allowing the model to ground its reasoning in intermediate tool outputs and self-correct during long-horizon tasks. To effectively train this complex interleaved policy, we introduce {Dr. GRPO}, a robust optimization objective that utilizes token-level loss aggregation with importance sampling to mitigate reward sparsity and high-variance credit assignment. Furthermore, we employ a dual-component reward system and dynamic curriculum via online rollout filtering to ensure structural integrity and sample efficiency. Extensive experiments reveal that our approach, despite being trained solely on math tasks, achieves state-of-the-art performance across diverse reasoning domains, demonstrating high token efficiency and strong generalization capabilities. 
\end{abstract}

\section{Introduction}
Large language models (LLMs) have achieved impressive performance in a wide range of reasoning and problem-solving tasks, especially when augmented with external tools such as code interpreters, calculators, and knowledge bases \citep{deepseek-ai-2025,qwen3technicalreport,team2025kimi}. These tool-augmented agents are capable of autonomously planning, invoking tools, and solving complex tasks that require multi-step reasoning \citep{gao2025survey,luo2025large,plaat2025agentic}. While prior studies have focused on supervised fine-tuning (SFT) and reinforcement learning (RL) to enhance tool usage, most approaches rely on domain-specific training or treat tool execution as a disjoint step from reasoning, limiting the agent's ability to generalize to unseen scenarios.

In this paper, we address a fundamental question: \textit{Can an LLM agent trained to use a code interpreter solely on mathematical problems generalize its reasoning and tool usage strategies to diverse, unseen domains?} To answer this question, we focus on a challenging setting where the agent learns tool invocation strategies via RL in the strict logic of the math domain and is subsequently evaluated on distinct, open-ended tasks. This setup enables us to systematically analyze cross-domain generalization and skill transfer. We observe that standard RL objectives (e.g., PPO or vanilla GRPO) often struggle in this context due to the \textit{linear} nature of traditional tool use and the severe reward sparsity inherent in long-horizon reasoning.

To overcome these limitations and enhance cross-domain transfer, we propose the \textbf{R}einforcement Learning for \textbf{I}nterleaved \textbf{T}ool \textbf{E}xecution (RITE) framework. Our approach is built on the insight that robust generalization stems from the reasoning structure rather than domain-specific knowledge.

\begin{tcolorbox}[notitle, rounded corners, colframe=darkgrey, colback=white, boxrule=2pt, boxsep=0pt, left=0.15cm, right=0.17cm, enhanced, shadow={2.5pt}{-2.5pt}{0pt}{opacity=5,mygrey},toprule=2pt, before skip=0.65em, after skip=0.75em 
  ]
\emph{
  {
    \centering 
  {
    \fontsize{9pt}{13.2pt}\selectfont 
 Learn to use tools rather than specific domain knowledge can truly enhance cross-domain generalization.
}  \\
  }
  }
\end{tcolorbox}

\begin{figure*}[t]
    \centering
    \includegraphics[width=0.85\textwidth]{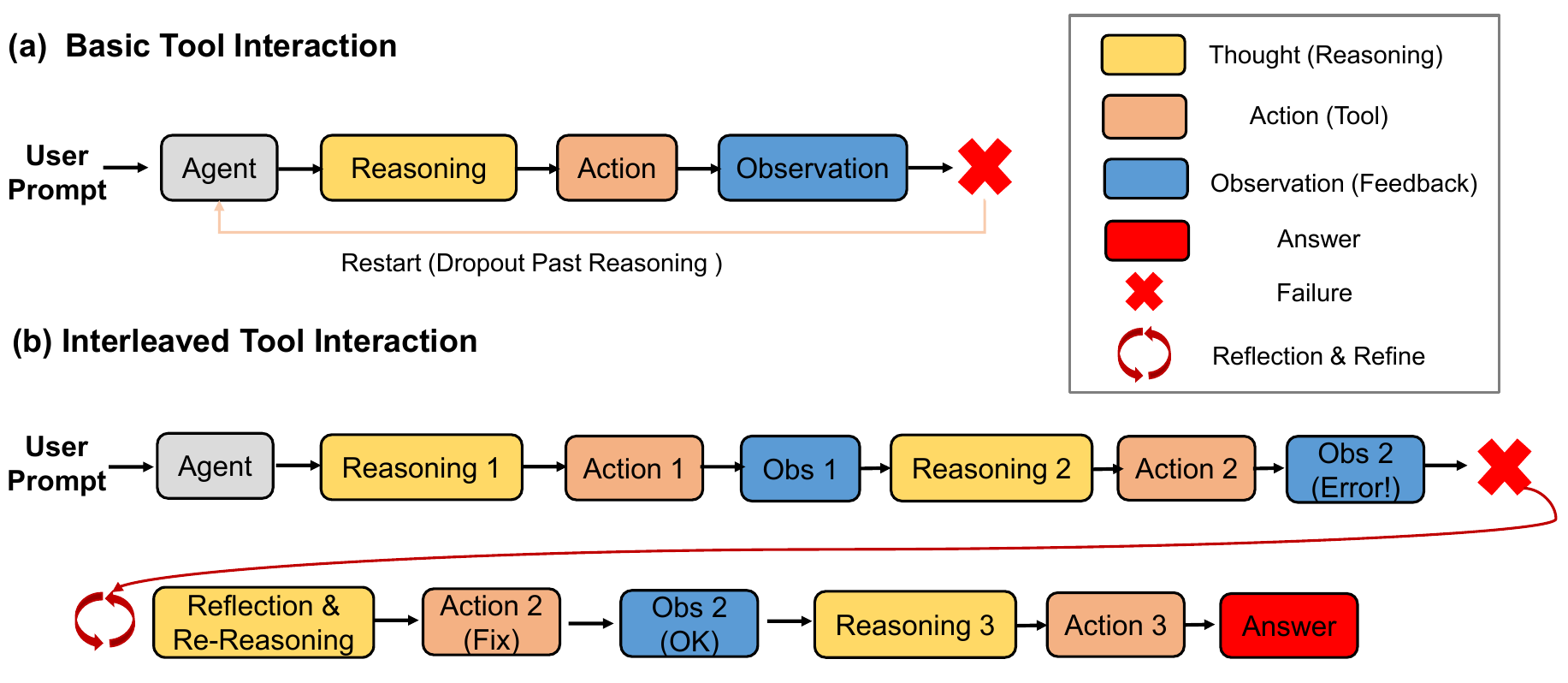}
    \caption{Interaction comparison between standard tool RL interactions (a) and our Interleaved Thinking process (b). While standard methods often discard intermediate states, our framework forms a continuous ``Plan-Action-Reflection'' loop, preserving reasoning history for error correction and robust generalization.}
    \label{fig:tool_rl_comparsion}
\end{figure*}

Specifically, our proposed RITE framework comprises three key technical contributions designed to stabilize training and foster domain-agnostic learning.
\textit{Tool-Integrated Interleaved Thinking:} We move beyond the separation of reasoning and execution by enforcing a ``Plan-Action-Reflection'' cycle as shown in  Figure \ref{fig:tool_rl_comparsion}. This alternating pattern ensures that the agent grounds every step in the latest tool observations, significantly reducing hallucination and error propagation when transferring to complex tasks.
    
 \textit{Token-Level Optimization:} Training long-context interleaved policies is unstable with trajectory-level rewards alone. We introduce Dr. GRPO, which incorporates token-level loss aggregation using Importance Sampling (IS) and Rejection Sampling (RS). This allows gradients to focus on critical decision points (e.g., tool invocation and reasoning transitions) rather than being diluted by long reasoning chains.
    
     \textit{Robust Training Curriculum:} To further align the agent with the interleaved structure, we implement a \textit{Dual-Component Reward} system (combining outcome correctness with structural formatting rewards) and a \textit{Dynamic Difficulty Adjustment} strategy that filters training samples based on the agent's pass rate, ensuring continuous learning within the Zone of Proximal Development.

Our contributions can be summarized as follows: First, we demonstrate that RL-based tool usage learned from mathematical tasks can be effectively transferred to diverse domains when the reasoning process is structured correctly.
Second, we introduce the Reinforcement Learning for Interleaved Tool Execution (RITE) framework with components specifically designed to enhance RL stability and cross-domain generalizability. Third, through extensive evaluations across multiple domains using diverse benchmarks, we achieve state-of-the-art performance, providing insights into the key factors driving successful skill migration and highlighting the transformative potential of Tool RL for LLM reasoning in cross-domain settings.

\section{Related Work}

\subsection{Tool-Integrated RL for LLM Reasoning}
Enhancing LLM reasoning with tools has achieved great attention due to its potential for expanding the knowledge boundary of models' generative support, thereby breaking the “invisible leash” that constrains purely text-based models \citep{lin2025understanding}. 
The code interpreter and search engine, as two representative tools, have been widely integrated into the natural-language reasoning process to individually boost the model's performance on mathematical \citep{feng2025retool,li2025torl} and knowledge-intensive reasoning tasks \citep{jin2025search,liu2025webexplorer,tongyidr}. Advanced techniques focus on addressing new challenges under more complex scenarios(e.g.multi-turn) from different perspectives, such as long-horizon planning \citep{gao2025beyond,singh2025fathom,erdogan2025plan}, memory management \citep{xu2025mem,yan2025memory,zhou2025mem1}, interaction efficiency \citep{wang2025otc,song2025r1}, multi-tool selection \cite{dong2025tool,zou2025autotool} and interleaved thinking structure \citep{xie2025interleaved,interleaved,liu2025deepseek}. However, neither of these works explores the generalizability of tool-integrated RL approaches across diverse reasoning tasks and domains.

\begin{figure*}
\centering
\setlength{\abovecaptionskip}{0cm}   
\setlength{\belowcaptionskip}{0cm}   
\subfigure[Performance]{
\includegraphics[width=0.3\linewidth]{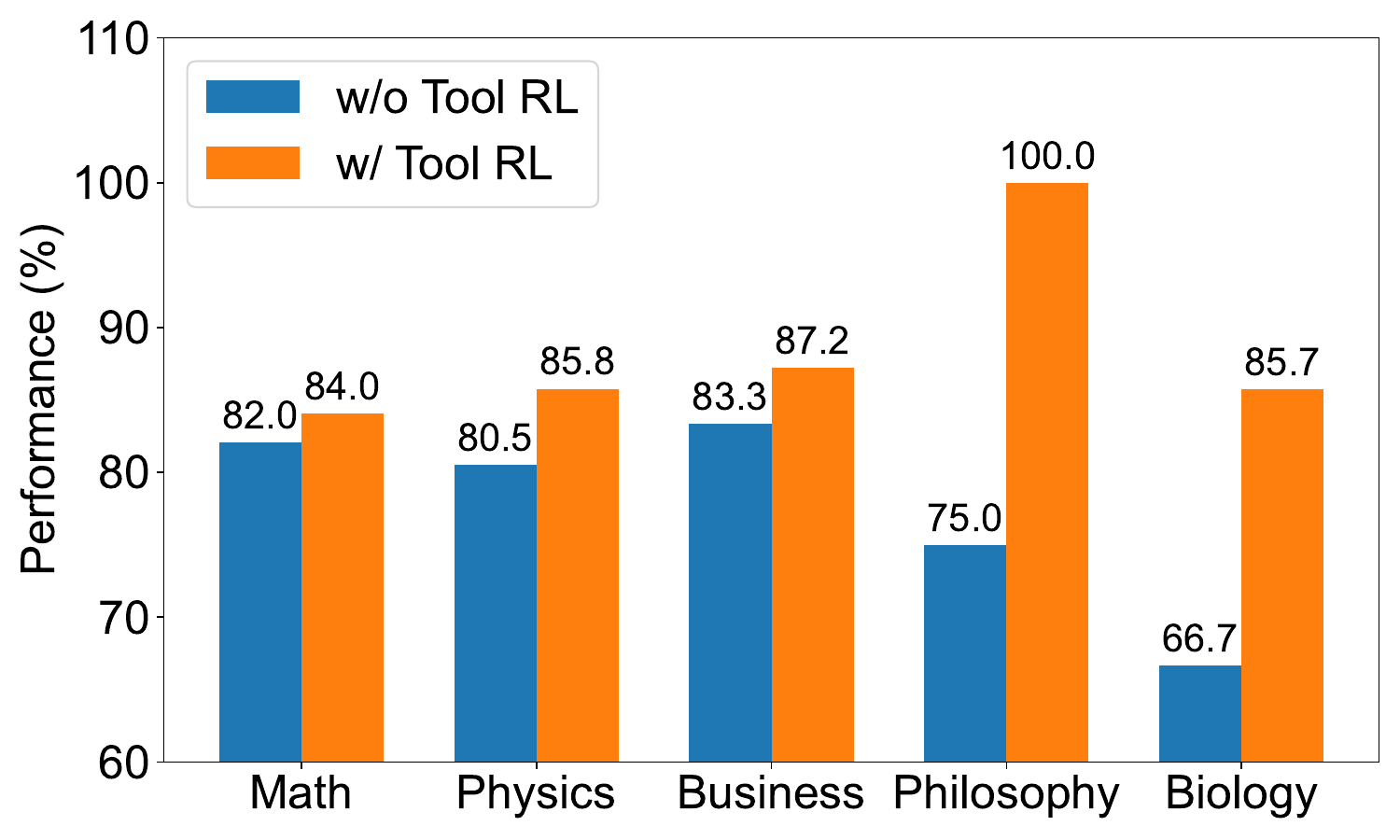}
}
\subfigure[Interaction Turn]{
\includegraphics[width=0.3\linewidth]{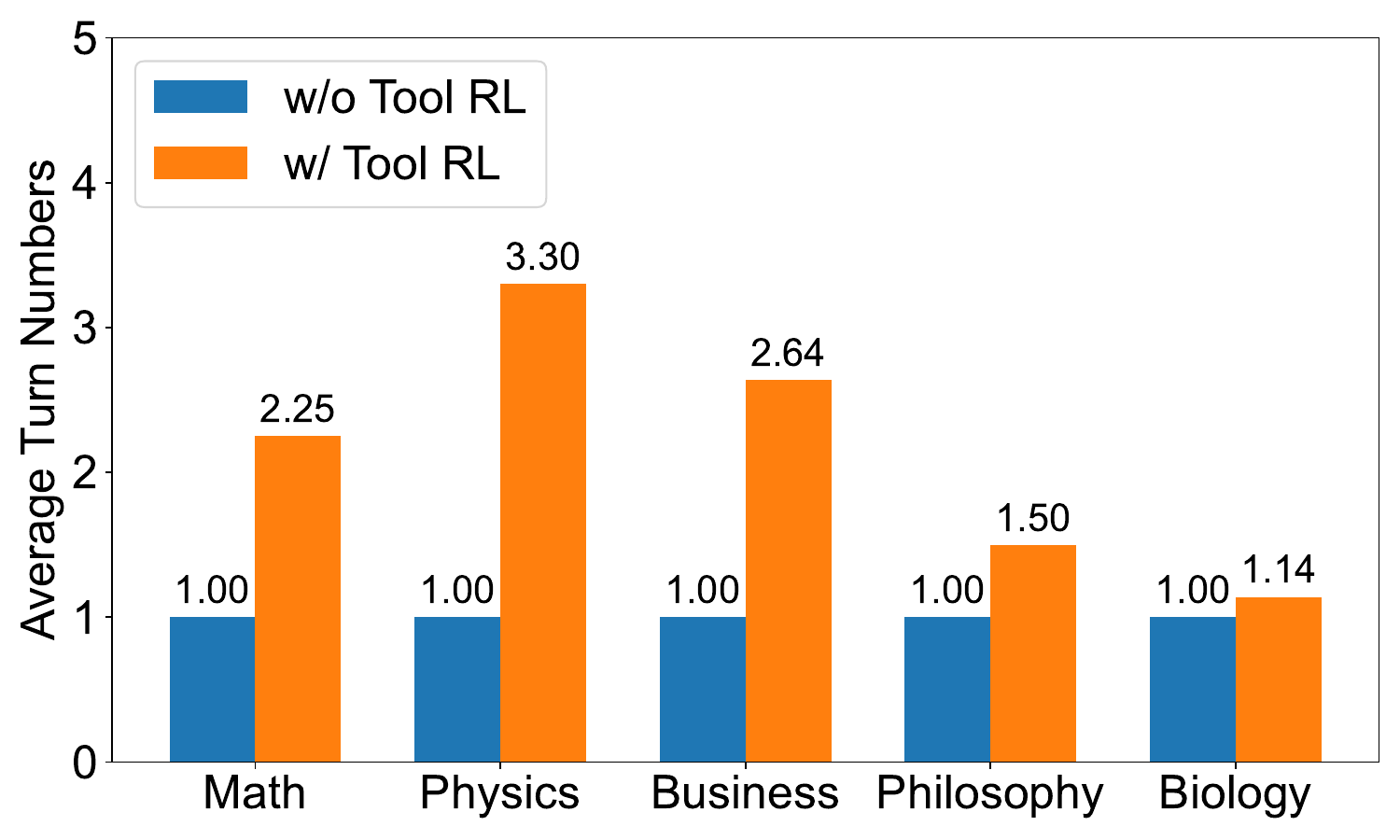}
}
\subfigure[Token Length]{
\includegraphics[width=0.3\linewidth]{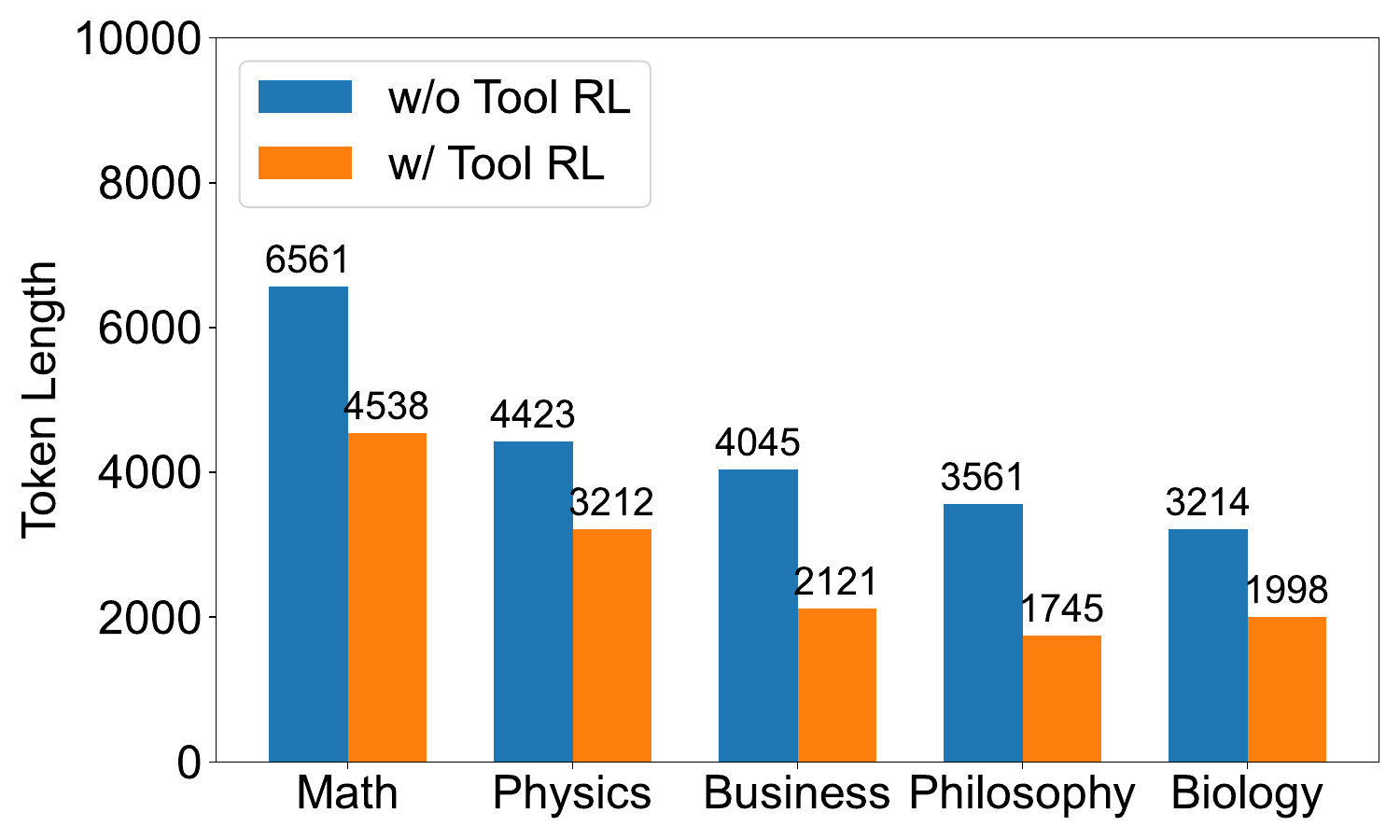}
}
\caption{Cross-domain comparisons including model performance, interaction turns, and token length for the output on Webinstruct, where we perform tool RL training on Qwen2.5-7B using the code-integrated math dataset.}
\label{motivation}
\end{figure*}

\subsection{Cross-Domain Reasoning for LLM}
Beyond mathematical and code domains, many efforts have been devoted to 
enhancing the cross-domain reasoning ability for LLMs \citep{cheng2025revisiting,li2025can}.  Some pioneering works focus on how to curate high-quality cross-domain datasets \citep{akter2025nemotron,ma2025general}, including both synthetic and real-world questions into RL training to improve generalization across diverse reasoning tasks. Moreover, compared with calculating reward solely depending on the model-based verifier \citep{xie2025capo}, the scaling and reliable reward signal from models' intrinsic mechanism have also been explored to broaden the reasoning boundary to general domains, assisted by Verifier-Free frameworks \citep{zhou2025reinforcing,yu2025rlpr,liu2025nover} and Rubrics \citep{su2025crossing,gunjal2025rubrics,huang2025reinforcement,zhou2025breaking}. The most similar to our topics are those that focus on the transferability of LLM reasoning \cite{huan2025does,zhou2025does,hu2025breaking,cheng2025revisiting}, but these works only 
clarify the generalization effect through single-domain training without tool invocation. Our work first systematically explores the cross-domain reasoning potential of Tool RL training.

\section{Uncover the Cross-Domain  Generalization Brought by Tool RL}

\subsection{Problem Formulation}
Our goal is to explore whether an LLM agent can generalize its tool-integrated reasoning performance on general domains, despite training exclusively on code-integrated math data $\mathcal{D}_{\text{CI}}$. Ideally, the ultimate target is to maximize expected performance on general domain queries as follows:

\begin{equation}
\max_{\theta} \mathbb{E}_{x \sim \mathcal{D}_{\text{Gen}}} \left[ \mathbb{E}_{y \sim \pi_{\theta}(\cdot \mid x,\text{tool})} \left[ R(y,x) \right] \right]
\end{equation}
where $\mathcal{D}_{\text{Gen}}$ represents the general domain query distribution and $R(y,x)$ is the reward function evaluating response $y$ for query $x$.

\subsection{Bridging the Domain Gap via Tools}

From the theoretical perspective, we advocate that we can
bridge the domain gap through tool-necessity optimization. By emphasizing high tool-necessity samples in $\mathcal{D}_{\text{CI}}$, we force the model to learn generalizable tool-usage reasoning patterns rather than domain-specific superficial features. The tool-necessity acts as a domain-invariant indicator of when tools are truly beneficial, enabling effective transfer to general domains. Thus, the formalization of the generalization objective to general domains is:
\begin{equation}
\max_{\theta} \mathbb{E}_{x \sim \mathcal{D}_{\text{gen}}} \left[ \mathbb{I}[\text{tool-needed}(x)] \cdot R(\pi_{\theta}(x,\text{tool}), x) \right]
\end{equation}
where $\mathbb{I}[\text{tool-needed}(x)]$ indicates queries in general domains that require tool assistance.

This approach ensures that even when trained on specialized math data, the model develops robust tool-integrated reasoning capabilities that generalize broadly beyond the training distribution.

\begin{figure*}[t]
\centering
  \includegraphics[width=0.8\textwidth]{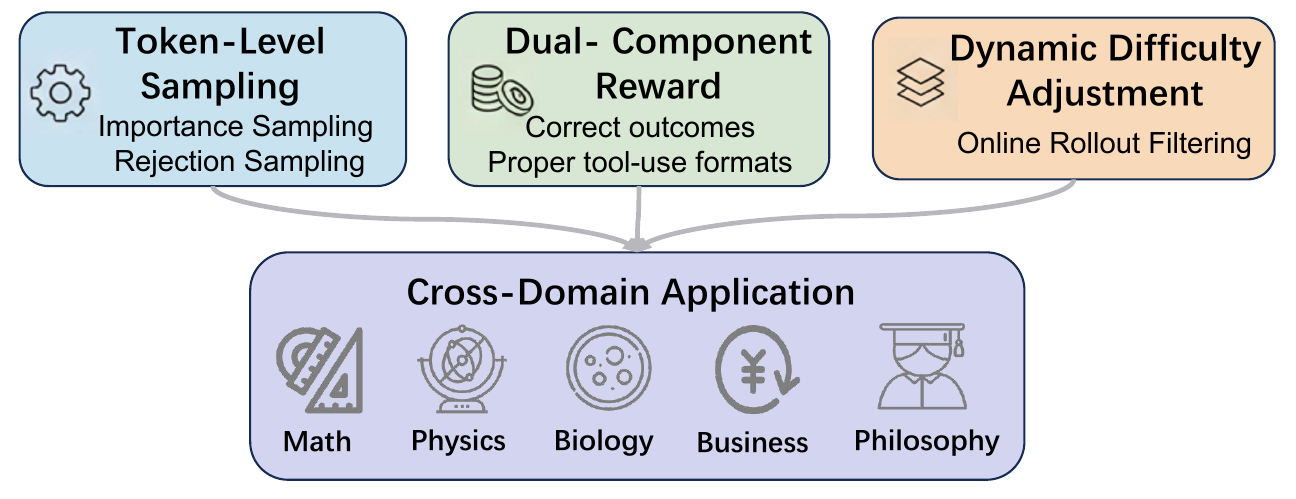}
  \caption{Illustrations of our proposed Reinforcement Learning for Interleaved Tool Execution (RITE) framework.}
  \label{method}
\end{figure*}
\subsection{Analysis of Cross-Domain Experiments}
We conduct exploration experiments to understand the generalization of tool-integrated RL approaches across diverse reasoning domains. As shown in Fig. \ref{motivation}, we perform the tool RL training of the Qwen2.5-7B on the code-integrated math data (as stated by Sec.\ref{sec_experiments}) and compare the model performance, the interaction turn numbers, and the token length of response across 5 different domains (Mathematics, Physics, Business, Philosophy, and Biology) on the Webinstruct \citep{ma2025generalreasoneradvancingllmreasoning} evaluation dataset.

We can observe that: (i) Comparing with and without tool RL training, integrating the tool call into the natural reasoning process can enhance model performance with more interaction turns and less token length; (ii) Despite training only on the code-integrated math data, the model performance and the number of interaction turns and token length for the general reasoning (out of domain, such as physics, biology and so on) output have the consistent trend with in-domain math dataset. This can verify that there exists some tool use patterns (orthogonal to domain knowledge) transferred from the code-integrated math domain to the general reasoning domain, which can enhance the cross-domain reasoning abilities during RL training.

\section{Reinforcement Learning for Interleaved Tool Execution}

\subsection{Tool-Integrated Interleaved Thinking}

To enable robust cross-domain generalization, we move beyond traditional tool-use paradigms (which often separate reasoning from execution) and introduce \textit{Tool-Integrated Interleaved Thinking}. Unlike standard methods where models perform comprehensive reasoning upfront or treat tool outputs as isolated events, Interleaved Thinking enforces an alternating cycle of \textit{inference} and \textit{tool invocation}.

\paragraph{The Plan-Action-Reflection Cycle}
As illustrated in Figure~\ref{fig:tool_rl_comparsion}, standard Tool RL often discards intermediate reasoning states or treats them linearly (Fig.~\ref{fig:tool_rl_comparsion}a). In contrast, our Interleaved Thinking framework (Fig.~\ref{fig:tool_rl_comparsion}b) preserves the entire reasoning history, forming a continuous ``Plan-Action-Reflection'' loop:
\begin{enumerate}
    \item \textbf{Plan (Thinking):} The model generates a dedicated thinking block to analyze the current state and formulate a hypothesis.
    \item \textbf{Action (Tool Call):} Based on the plan, the model invokes a tool (e.g., a code interpreter).
    \item \textbf{Reflection (Observation Integration):} Crucially, the model retains the reasoning context and the tool output to self-reflect, correcting deviations before the next step.
\end{enumerate}

This alternating pattern is vital for long-horizon tasks. By grounding each step in the latest tool execution results rather than initial assumptions, the model reduces error propagation and hallucination—key factors when transferring skills from math domains to complex, open-ended tasks.

\subsection{Robust Optimization Objective for Long-Horizon Tool Reasoning}

Training the Interleaved Thinking policy introduces unique challenges not adequately addressed by standard objectives like  GRPO. The multi-turn nature of the "Plan-Action-Reflection" cycle often involves extremely long contexts (up to 62K tokens) and deep interaction horizons (up to 200 turns), leading to:
(i) \textbf{severe reward sparsity}, where success is only observable at the end of a long reasoning trajectory; and
(ii) \textbf{high-variance credit assignment}, where early planning errors dominate the final outcome but receive weak learning signals.

To address these issues, we propose a robust optimization objective tailored for Interleaved Thinking, built upon Group Relative Policy Optimization (GRPO) \cite{shao2024deepseekmath} with a novel token-level loss aggregation strategy.

\textit{Group Relative Policy Optimization}.
For a given problem instance $q$, we sample a group of $K$ trajectories $\{\tau_i\}_{i=1}^K$ generated via the interleaved process. Each trajectory $\tau_i$ yields a final scalar reward $R_i$. Instead of relying on absolute rewards, GRPO computes a \emph{relative advantage} within the group:
\begin{equation}
A_i = R_i - \frac{1}{K} \sum_{j=1}^{K} R_j.
\end{equation}
This relative formulation automatically normalizes reward scales across problems of varying difficulty, mitigating training instability caused by the heterogeneous task distributions typical in cross-domain transfer.

\textit{Dr. GRPO: Token-Level Loss Aggregation}. While GRPO stabilizes optimization at the trajectory level, directly aggregating loss over all tokens remains brittle for long interleaved sequences. We therefore utilize Dr. GRPO, which introduces token-level loss aggregation to improve gradient fidelity \cite{liu2025understanding}.

Let $r_{i,t}$ denote the policy ratio at token position $t$ of trajectory $\tau_i$:
\begin{equation}
r_{i,t} = \frac{\pi_{\theta}(a_{i,t} \mid s_{i,t})}{\pi_{\theta_{\text{old}}}(a_{i,t} \mid s_{i,t})}.
\end{equation}

The Dr. GRPO objective is defined as:
\begin{equation}
\begin{aligned}
\mathcal{L}_{\text{Dr-GRPO}} &= \sum_{t=1}^{T} w_t \cdot \mathbb{E}_i \Big[ - \min \big( r_{i,t} A_i, \\
&\quad \text{clip}(r_{i,t}, 1-\epsilon, 1+\epsilon) A_i \big) \Big],
\end{aligned}
\end{equation}
where $w_t$ is a token-level aggregation weight. This formulation allows gradients to focus on tokens that are causally important for the "Plan-Action" decisions, rather than being diluted by long spans of low-impact text.

\subsection{Token-Level Importance and Rejection Sampling}
We construct the token-level weight $w_t$ using a combination of \textit{Token-level Importance Sampling (IS)} and \textit{Token-level Rejection Sampling (RS)}:
\begin{equation}
w_t = \alpha \cdot \text{IS}_t + (1 - \alpha) \cdot \mathbb{I}[\text{RS}_t],
\end{equation}
where $\text{IS}_t$ prioritizes tokens involved in tool invocation and reasoning transitions (the "thinking" tags), while $\text{RS}_t$ downweights stylistic or repetitive tokens. This decomposition is essential for reinforcing the structural integrity of the interleaved thinking process.

\subsection{Dual-Component Reward System}

To guide the model toward effective Interleaved Thinking while ensuring robustness across domain shifts, we implement a dual-component reward system. The overall reward function is:
\begin{equation}
R = R_{\text{outcome}} + R_{\text{format}},
\end{equation}
with $R \in \{-2,-1,0,1,2\}$.

\subsubsection{Outcome Reward}
The Outcome Reward evaluates the correctness of the final answer, targeting domain-invariant problem-solving skills. Let $\hat{a}$ denote the model’s output and $a^\ast$ the ground-truth solution:
\begin{equation}
R_{\text{outcome}} = 
\begin{cases}
+1, & \text{if } \hat{a} \equiv a^\ast,\\
-1, & \text{otherwise}.
\end{cases}
\end{equation}
This reward encourages abstract reasoning (e.g., mapping problems to computable steps) regardless of the domain.
\begin{figure}[t]
    \centering
    \includegraphics[width=0.47\textwidth]{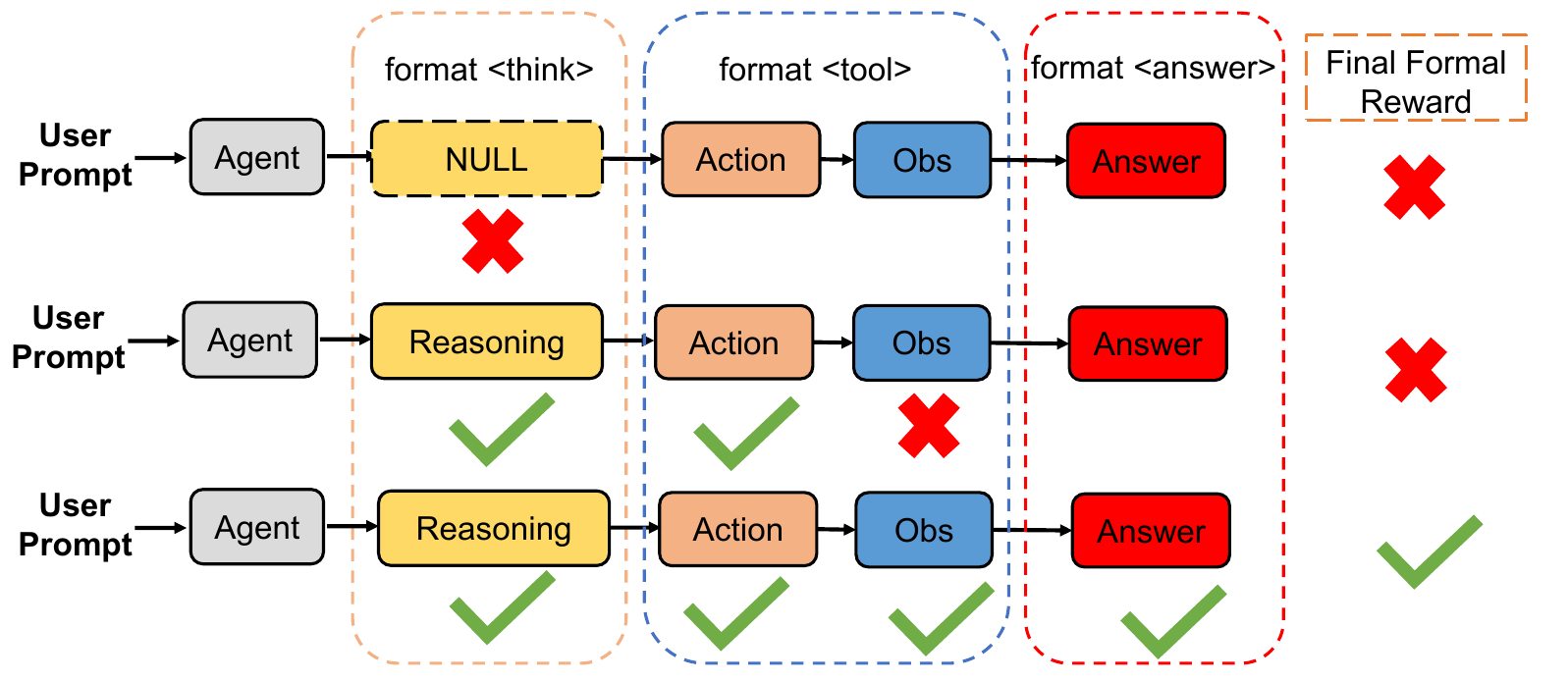}
    \caption{Illustration of format reward. The agent must maintain specific Interleaved Thinking structure (Plan-Action-Reflection) to receive positive reinforcement.}
    \label{fig:format_reward}
\end{figure}
\subsubsection{Format Reward}
The Format Reward is critical for Interleaved Thinking. It incentivizes the maintenance of the correct "thinking" and "tool use" structure, ensuring the agent does not degenerate into unstructured generation. Our format reward can be defined:
\begin{equation}
R_{\text{format}} =
\begin{cases}
+1 & \text{valid interleaved structure}, \\
0  & \text{minor formatting issues}, \\
-1 & \text{broken thinking/tool cycle}.
\end{cases}
\end{equation}
As shown in Figure~\ref{fig:format_reward}, only when the interaction includes necessary content with correct interleaved format can the agent achieve a positive reward.

\subsection{Dynamic Difficulty Adjustment via Online Filtering}

To efficiently train Interleaved Thinking on complex tasks, we address the sample efficiency problem caused by static datasets. We utilize \textit{Online Rollout Filtering}, a dynamic curriculum strategy.

For each prompt $q$, we maintain a running estimate of its pass rate:
\begin{equation}
\text{PassRate}(q) = \frac{\text{\# successful rollouts}}{\text{\# total rollouts}}.
\end{equation}

Prompts are included in training only if $\text{PassRate}(q) \in [0.1, 0.9]$. This ensures the agent continuously trains on problems within its \emph{Zone of Proximal Development}, where the interleaved reasoning strategy is actively being refined, rather than on trivial or impossible tasks.




\section{Experiment}\label{sec_experiments}

\begin{table*}
    \centering
    \resizebox{\linewidth}{!}{
    \setlength\tabcolsep{2.2pt}
    \begin{tabular}{lcc|ccc|cccc|ccc} 
    \toprule
     \multirow{2}{*}{\textbf{Model}} & \multirow{2}{*}{\textbf{TIR}} & \multirow{2}{*}{\textbf{Zero-RL}}    & \textbf{GPQA} & \textbf{TheoremQA} & \textbf{WebInst.} & \textbf{MATH-500} & \textbf{HMMT 25} & \textbf{AIME 24} & \textbf{AIME 25} & \textbf{General} & \textbf{All}\\ 
 & &  & Avg@4 & Avg@2 & Avg@2 & Avg@2 & Avg@2 & Avg@16 & Avg@16 & - & - & \\

\midrule
 \multicolumn{13}{c}{\textbf{7B Size Models}} \\ 
 
\midrule
Qwen2.5-7B & \ding{55} & -- & 32.4	 &  41.4  &  60.4 & 51.9 & 0.0 & 3.2 & 1.1  & 44.7 & 27.2 \\ 
Qwen2.5-7B-TIR & \ding{51} & -- &  28.5		 & 35.2  & 52.3  & 18.0 &1.9 & 1.7 & 0.6     & 38.7 & 19.7 \\ 
\midrule
ToRL-7B & \ding{51} & \ding{51} & 35.8	  & 48.6	  & 68.2  & 82.2 &26.3 & 40.2  &  27.9   & 50.9 & 47.0 \\
ZeroTIR-7B & \ding{51} & \ding{51} & 34.2	  & 46.8	  &  65.7 & 80.2 &22.5 &39.6 &25.0   & 48.9 & 44.9 \\
SimpleTIR-7B & \ding{51} & \ding{51} & \textbf{36.2}	  & 49.5	  & 70.7  & \textbf{88.4} &29.7 & \textbf{50.5} & 30.9      & 52.1 & 50.8 \\  
\rowcolor[HTML]{D7E8E8}
 RITE-7B(Ours) & \ding{51} & \ding{51} &  35.6	 & \textbf{50.3}  &  \textbf{73.6} & 86.0 & \textbf{31.1} & \textbf{50.5}&  \textbf{36.7}   & \textbf{53.2} & \textbf{52.0} \\
\midrule
 \multicolumn{13}{c}{\textbf{32B Size Models}} \\ 
 
\midrule
Qwen2.5-32B & \ding{55} & -- & 38.6	  &  48.2	 & 68.5  & 43.1& 0.2 & 4.2 &1.6      & 51.8 & 29.2 \\ 
Qwen2.5-32B-TIR & \ding{51} & -- & 35.1	   & 42.8	 &61.9    & 37.0 &5.2 &7.1  &5.0   & 46.6 & 27.7 \\ 
\midrule
ReTool-32B & \ding{51} & \ding{55} & 42.3	 &54.7	   &76.8  &  93.2 & 36.2 &  67.0 & 49.3  & 57.9 & 59.9 \\
SimpleTIR-32B & \ding{51} & \ding{51} &43.5	   &56.3	 &78.1   & 92.9 & 34.6  &  59.9 & 49.2  & 59.3 & 59.2 \\ 
\rowcolor[HTML]{D7E8E8}
 RITE-32B(Ours) & \ding{51} & \ding{51}   & \textbf{46.4}	& \textbf{57.4}	& \textbf{82.3} & \textbf{93.8} & \textbf{39.5} & \textbf{71.3} & \textbf{56.7} & \textbf{62.0} & \textbf{63.9}\\
    \bottomrule
    \end{tabular}
    } 
    \caption{Main results on seven reasoning benchmarks. ``TIR'' denotes Tool-Integrated Reasoning. {General} is the average of GPQA, TheoremQA, and WebInst. Notably, RITE is trained \textit{only} on math tasks but achieves best on general domains.}
    \label{tab:main_exp}
\end{table*}

\begin{figure*}[t]
    \centering
    \begin{minipage}{0.95\linewidth}
        \centering
        \begin{minipage}{0.45\linewidth}
            \centering
            \includegraphics[width=\linewidth]{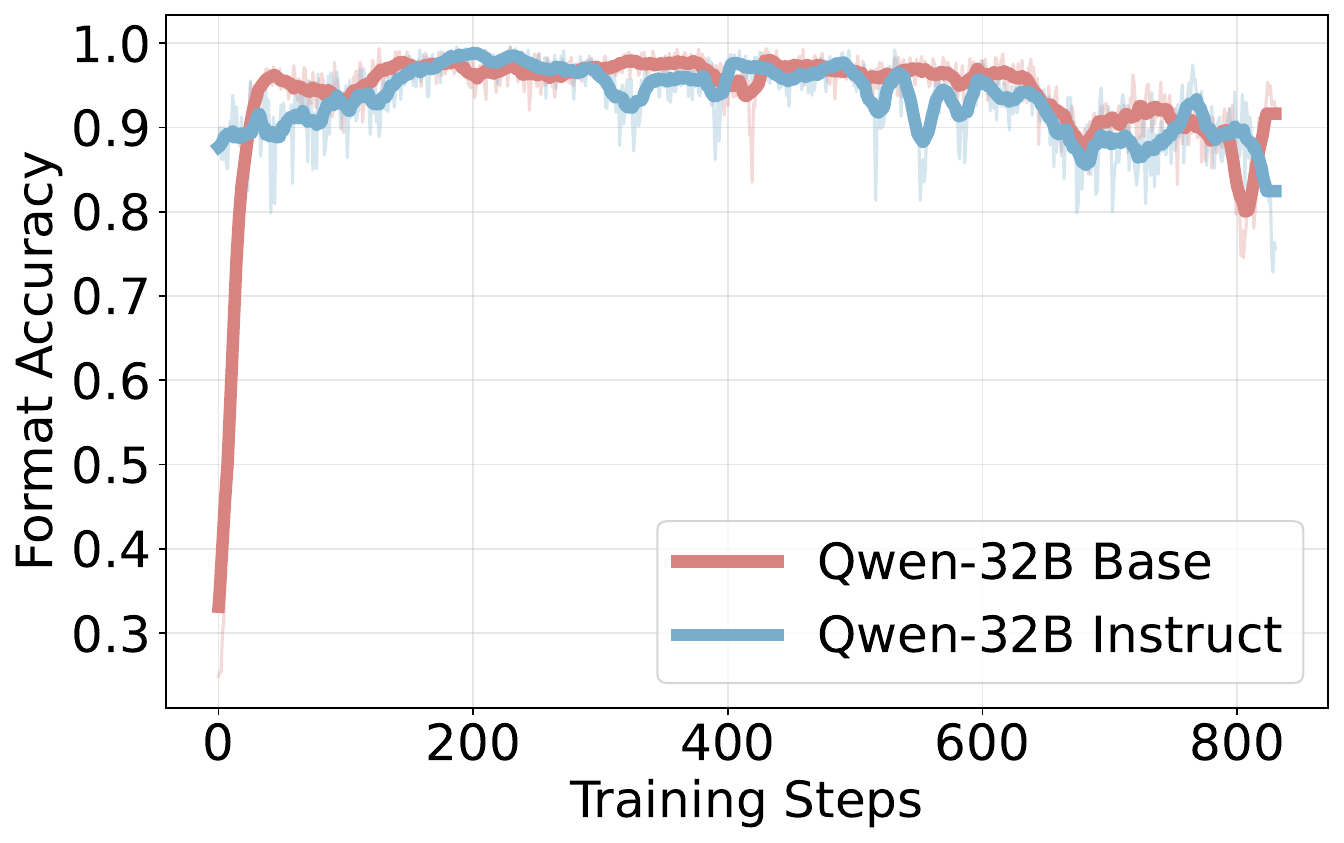}
            \caption*{Training Steps vs Format Accuracy}
        \end{minipage}
        \hspace{0.08\linewidth}
        \begin{minipage}{0.45\linewidth}
            \centering
            \includegraphics[width=\linewidth]{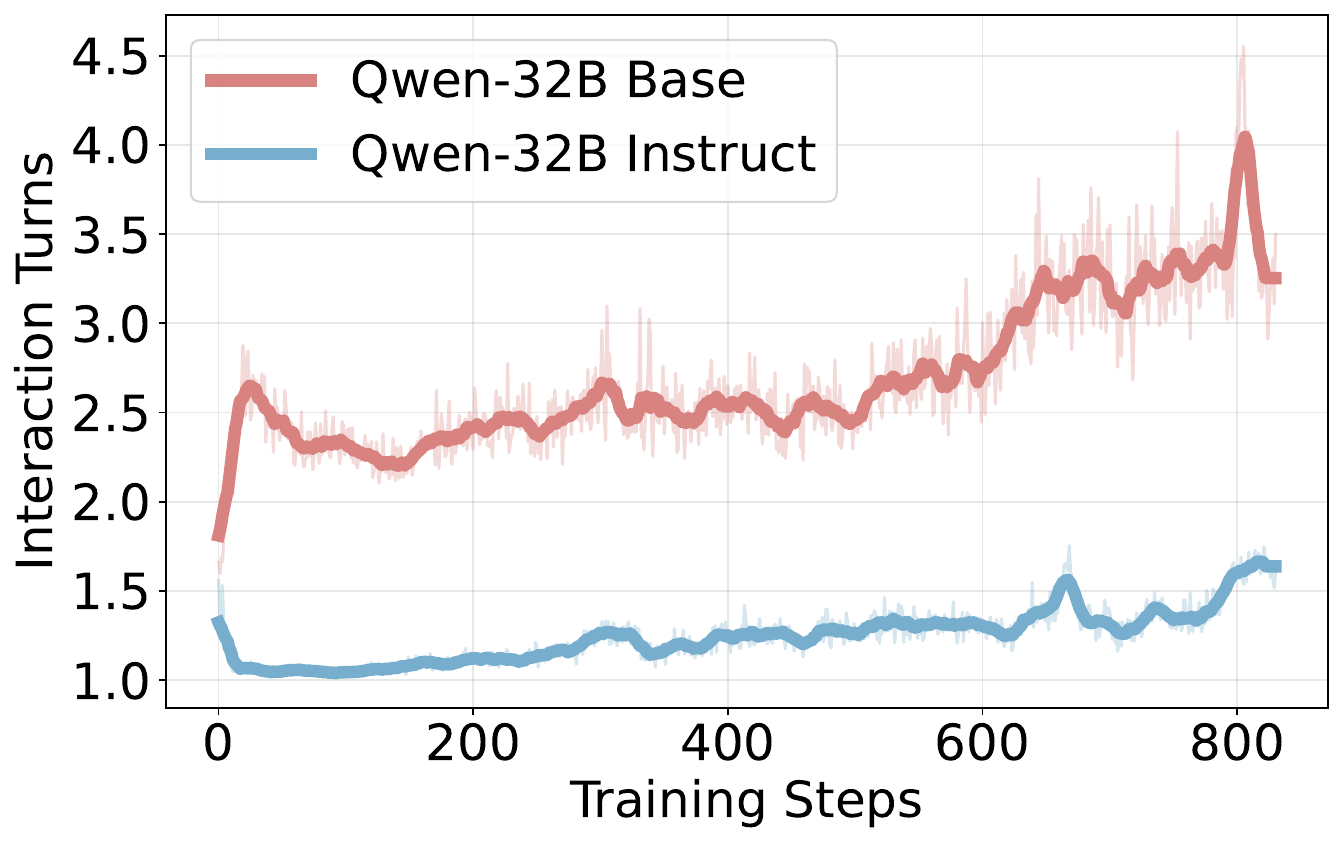}
            \caption*{Training Steps vs Interaction Turns}
        \end{minipage}
    \end{minipage}
    \caption{Training progress of format accuracy and interaction turns.}
    \label{fig:train_format_interaction}
\end{figure*}

\begin{figure*}
\centering
\subfigure{
\includegraphics[width=0.45\linewidth]{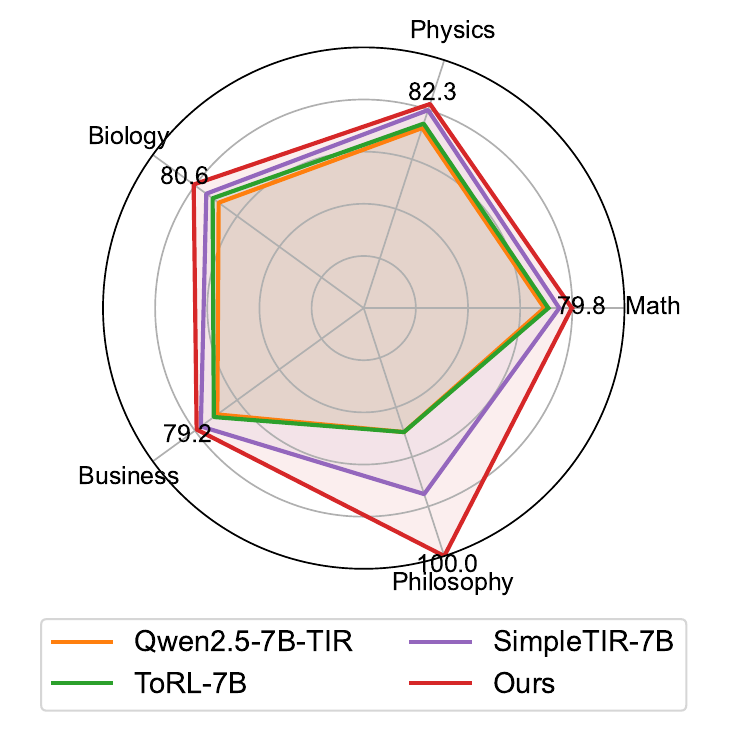}
}
\subfigure{
\includegraphics[width=0.45\linewidth]{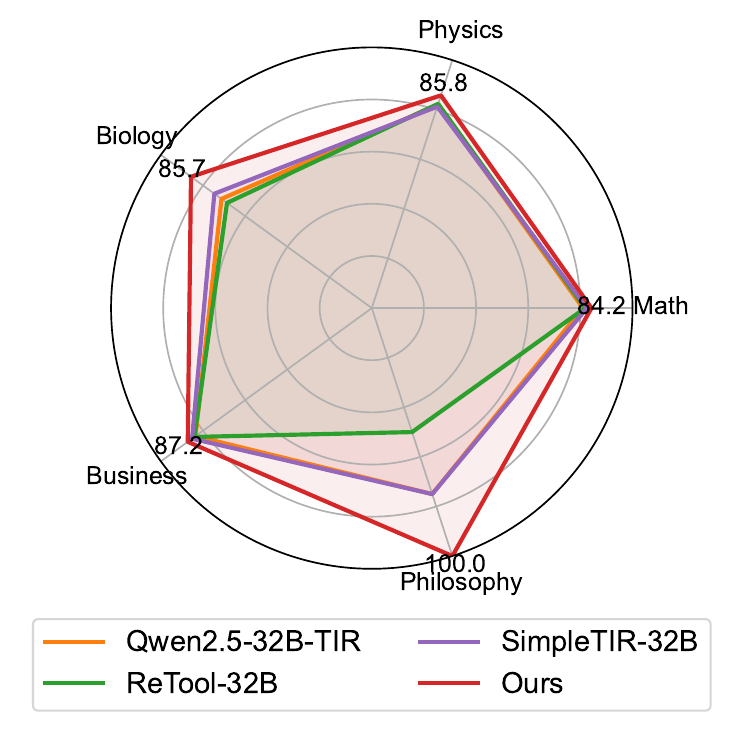}
}
\caption{Cross-domain performance comparison between different Tool RL methods on Webinstruct.}
\label{results_domain}
\end{figure*}

We design our experiments to answer the core research question: \textit{Can an agent trained exclusively on mathematical tool-use tasks generalize its reasoning capabilities to diverse, open-ended domains?} To this end, we train our models solely on code-integrated math datasets and evaluate them on a broad spectrum of reasoning benchmarks, including science, logic, and general knowledge.


\subsection{Experimental Setup}

\paragraph{Datasets and Environment.}
Our training data is strictly confined to the mathematical domain. We utilize the \textbf{Math3-5} dataset~\citep{zeng2025simplerl} and the math-subset of \textbf{DeepScaler}~\citep{deepscaler2025}, filtering for samples solvable via Python code. The environment consists of a sandboxed Python interpreter. 
For evaluation, we classify benchmarks into two categories:
\textit{In-Domain (Math):} MATH-500, HMMT 25, AIME 24, and AIME 25.
 \textit{Out-of-Domain (General):} GPQA (Graduate-Level Science), TheoremQA (STEM), and the held-out test set of WebInstruct (Multidisciplinary).

\paragraph{Implementation Details.}
We train two variants of RITE based on {Qwen2.5-7B} and {Qwen2.5-32B}. Unlike standard approaches that require Supervised Fine-Tuning (SFT) warm-up, we employ a {Cold-Start RL} strategy (Zero-RL), initializing directly from the base or instruct checkpoints without specific tool-tuning.
We use the {Dr. GRPO} objective with a group size of $G=16$. The token-level importance weight $\alpha$ is set to $0.8$. We implement the {Dynamic Difficulty Adjustment} by filtering prompts with pass rates outside $[0.1, 0.9]$. The maximum context length expands from 16k to 24k during training to accommodate the growing depth of the Interleaved Thinking cycles.



\subsection{Baseline Models}

We evaluate our approach against a comprehensive set of baseline models across two parameter scales to demonstrate its effectiveness.

\paragraph{7B Parameter Baselines}: \textbf{Qwen2.5-7B}, the  instruct model; \textbf{Qwen2.5-7B-TIR}, a tool-integrated reasoning with instruct model; \textbf{ToRL-7B}~\citep{li2025torl}, which applies domain-specific reinforcement learning to mathematical tool usage; \textbf{ZeroTIR-7B}~\citep{mai2025agent}, implementing zero-shot tool integration from base models; and \textbf{SimpleTIR-7B} \citep{xue2025simpletir}, a single-domain tool RL approaches.

\paragraph{32B Parameter Baselines}: \textbf{Qwen2.5-32B}, the instruct model; \textbf{Qwen2.5-32B-TIR}, a tool-integrated reasoning with instruct model; \textbf{ReTool-32B}~\citep{feng2025retool}, which employs cold-start supervised fine-tuning followed by domain-specific RL, and \textbf{SimpleTIR-32B} \citep{xue2025simpletir}, a scaled single-domain tool RL approaches.

These baselines enable systematic comparison across different training paradigms: instruct models without tool training, supervised fine-tuning approaches, domain-specific RL methods, and our proposed Reinforcement Learning for Interleaved Tool Execution (RITE) framework.

\subsection{Main Results}

Table~\ref{tab:main_exp} presents the comprehensive evaluation results. We have the following observations.

\paragraph{RITE achieves state-of-the-art results across both 7B and 32B model scales.} The 7B variant attains 36.7\% on AIME 25, outperforming prior tool-augmented baselines, and demonstrates strong cross-domain generalization with 53.2\% on averaged general reasoning benchmarks. Scaling to 32B parameters yields further gains: 71.3\% on AIME 24, 56.7\% on AIME 25, and 62.0\% on the general set, significantly surpassing comparable baselines. These results validate the framework's scalability and highlight the contribution of its standardized interface, dual-component reward, and structured prompting to superior performance in both mathematical and general reasoning tasks.

\paragraph{Strong Generalization from Math to Open Domains.}
The most significant finding is RITE's performance on \textbf{WebInstruct} and \textbf{GPQA}, benchmarks unrelated to the math training data.
RITE-7B achieves \textbf{73.6\%} on WebInstruct, surpassing the math-trained ToRL-7B by 5.4\%. Similarly, RITE-32B sets a new state-of-the-art with \textbf{82.3\%}, significantly outperforming ReTool-32B (76.8\%). This validates our hypothesis: \textit{the ``Plan-Action-Reflection'' cycle learned in math is a transferable cognitive structure.} Unlike baselines that overfit to math-specific templates, RITE's interleaved thinking allows the agent to break down general problems (e.g., physics or business logic) into executable sub-steps.

\paragraph{Robustness in Long-Horizon Math Tasks.}
On complex benchmarks like \textbf{AIME 25}, which requires deep multi-step reasoning, RITE shows a decisive advantage. RITE-7B achieves \textbf{36.7\%}, a relative improvement of \textbf{18.7\%} over SimpleTIR-7B. This improvement stems from {Dr. GRPO}, which effectively assigns credit in long trajectories where standard GRPO often fails due to reward sparsity.

\subsection{Analysis of Cross-Domain Transfer}

To visualize the transferability of the learned policy, as shown in Figure~\ref{results_domain}, we decompose performance on WebInstruct into distinct sub-domains: Math, Physics, Business, Philosophy, and Biology.

RITE (represented by the red area) encompasses the baselines in almost all dimensions. Notably, the performance gap is widest in \textbf{Physics} and \textbf{Business}. These domains share a structural similarity with mathematics—they require rigorous logic and step-by-step verification—but differ in vocabulary and knowledge retrieval. The \textit{Plan-Action-Reflection} mechanism allows the model to handle this shift: the "Plan" step grounds the problem in the current context, while the "Action" (tool use) offloads computation, regardless of whether the variables represent quantum states or financial assets.

\subsection{Training Dynamics}

Figure~\ref{fig:train_format_interaction} visualizes the evolution of the agent during training.
\textbf{Structural Adaptation:} The agent, starting from a base model with zero prior tool knowledge, achieves near-perfect format compliance (Interleaved Structure) within the first 50 steps. This rapid adaptation is driven by the \textbf{Format Component} of our Dual-Component Reward.
\textbf{Emergent Reasoning Depth:} As training progresses, the average interaction turns increase from $\sim 1.6$ to $\sim 4.0$. Unlike the instruct-tuned baseline which stagnates at lower interaction depths, the RITE agent learns to verify and self-correct, utilizing more turns to solve harder problems. This confirms that our \textbf{Dynamic Curriculum} successfully pushes the model to explore its Zone of Proximal Development.

\begin{table}[t]
    \centering
    \caption{Ablation study on AIME24 (In-Domain) and WebInstruct (Out-of-Domain).}
    \begin{tabular}{lcc}
    \toprule
    Method & AIME24 & WebInst. \\
    \midrule
    \textbf{RITE (Full)} & \textbf{71.3} & \textbf{82.3} \\
    \midrule
    w/o Token Opt. & 56.7 & 61.9 \\
    w/o Dual Reward & 67.0 & 76.8 \\
    w/o Interleaved & 59.9 & 78.1 \\
    \bottomrule
    \end{tabular}
    \label{tab:ablation}
\end{table}
\subsection{Ablation Study}

As shown in Table~\ref{tab:ablation}, we validate the contribution of each component in RITE using the 32B model. 

\paragraph{Impact of Token-Level Optimization (w/o Token Opt.).} Removing token-level optimization (replacing Dr. GRPO with standard GRPO and removing importance/rejection sampling) causes the most severe degradation, with AIME24 performance dropping by {14.6\%}. This highlights that for long interleaved reasoning chains, token-level optimization is critical to stabilize gradients and mitigate the high variance of credit assignment.

\paragraph{Impact of Dual-Component Reward (w/o Dual Reward).} Removing the format-aware reward leads to a significant drop in WebInstruct (-5.5\%). Without explicit structural reinforcement, the model struggles to maintain the ``Plan-Action-Reflection'' loop in open-ended domains, often reverting to unstructured text generation that fails to leverage tools effectively.

\paragraph{Impact of Interleaved Thinking (w/o Interleaved).} We trained a variant where tool execution is treated linearly (Plan $\to$ Code $\to$ Result) without the enforced reflection cycle. This led to a 4.2\% drop on WebInstruct. The lack of an explicit reflection step increases error propagation, proving that the cyclical structure is the key engine for cross-domain generalization.

\section{Conclusion}
We systematically examine the cross-domain generalization of large language model (LLM) agents equipped with a code interpreter, trained solely via reinforcement learning (RL) on mathematical problem-solving tasks. Our results show that RL-trained tool use in one domain transfers effectively to diverse reasoning tasks, demonstrating strong performance and high token efficiency.
To enable such transfer, we propose the Reinforcement Learning for Interleaved Tool Execution (RITE) framework, featuring a standardized tool interface, dual-component reward system, and XML-based prompt template to promote domain-agnostic learning. Extensive experiments demonstrate the effectiveness of our proposed RITE.


\section{Limitations}

Despite the promising results, several limitations remain in our current study:

\begin{itemize}
    \item \textbf{Restricted Tool Diversity:} Our experiments primarily focus on a single type of tool (code interpreter). The generalization performance across domains involving fundamentally different tools (e.g., knowledge base retrieval, image processing) remains to be explored.
    \item \textbf{Domain Shift Extremes:} While our benchmarks cover a variety of reasoning domains, they do not encompass highly specialized or adversarial domains where domain-specific knowledge or tool customization may be indispensable.
    \item \textbf{Reward Engineering:} The dual-component reward system relies on carefully designed heuristics and domain-agnostic abstraction, which may require manual tuning for new tasks or tools.
    \item \textbf{Scalability and Efficiency:} The framework assumes access to sufficient computational resources for RL training and evaluation. Scaling to more complex domains or larger toolsets may introduce additional efficiency and stability challenges.
\end{itemize}

Future work will address these limitations by extending the framework to support a broader range of tools, exploring more extreme domain shifts and improving scalability and prompt flexibility. We believe these directions will further advance the generalization of tool-augmented LLM agents.

\section{Ethical Considerations}

As large language models (LLMs) increasingly integrate with external tools and autonomous reasoning capabilities, it is imperative to address the ethical implications of their development and deployment. Our research on Reinforcement Learning for Interleaved Tool Execution (RITE) aims to enhance the reasoning and problem-solving abilities of AI agents; however, we acknowledge several critical considerations regarding safety, misuse, and bias.

\begin{itemize}
 \item \textbf{Safety and Code Execution Risks}
A core component of our framework is the integration of a code interpreter tool. While this significantly boosts mathematical and scientific reasoning, it introduces security risks associated with arbitrary code execution. To mitigate these risks during our experiments, all tool interactions were confined to a strictly sandboxed, non-networked Python environment with rigorous resource limits (CPU, memory, and timeout constraints). We strongly advise that any deployment of the models or frameworks released in this study must implement similar isolation mechanisms (e.g., Docker containers, gVisor) to prevent potential malicious exploitation, such as unauthorized file system access or denial-of-service attacks.
 \item \textbf{Dual-Use and Potential Misuse}
The enhanced reasoning and planning capabilities demonstrated by our agents—particularly their ability to generalize from mathematical tasks to broader domains—raise concerns regarding dual-use. While our primary goal is to advance scientific research and educational support, agents capable of complex multi-step planning and code generation could potentially be misused by malicious actors to automate cyber-attacks or generate obfuscated malware. We emphasize that the release of our models is intended for research purposes, and we encourage the community to develop robust safety guardrails and monitoring systems alongside capability improvements.
\item \textbf{Data Usage and Compliance}
Our training and evaluation processes utilized publicly available datasets (e.g., MATH, AIME, GPQA, WebInstruct). We have adhered to the licensing terms of these datasets and ensured that no personally identifiable information (PII) was processed or generated during our experiments. Our proposed data synthesis and filtering pipelines rely solely on open-source data and model-generated content, avoiding the use of private or proprietary user data.
\item \textbf{Environmental Impact}
Reinforcement learning, particularly with iterative rollout and rejection sampling, is computationally intensive. To minimize our carbon footprint, we employed efficiency-oriented techniques such as Online Rollout Filtering and dynamic context expansion, which prevent the waste of computational resources on samples that provide low information gain. 
\end{itemize}


\bibliography{custom}


\appendix

\section{Implementation Details}  
\subsection{Standardized Tool Interface}

We introduce a standardized tool interface as a critical component of our Reinforcement Learning for Interleaved Tool Execution framework, centered around a specialized \texttt{answer} tool. Unlike traditional tool-calling approaches that primarily focus on interacting with external APIs, our interface enforces a unified response format and serves as an explicit termination signal for the model’s reasoning process, abstracting domain-specific nuances to promote transferable invocation patterns.

Formally, the tool is defined as a function:
\begin{equation}
\mathcal{T}_{\text{answer}}: \mathcal{X} \mapsto \mathcal{Y},
\end{equation}
where $\mathcal{X}$ is the internal reasoning trace and $\mathcal{Y}$ is the final output. The output must satisfy the constraint
\begin{equation}
y \in \mathcal{Y}, \quad y = \texttt{\textbackslash boxed\{answer\}},
\end{equation}
with $a$ being the predicted solution content.

This design yields several advantages that facilitate cross-domain transfer:
\begin{itemize}
    \item \textbf{Format Standardization}: 
    Enforcing final answers in \verb|\boxed{}| ensures consistent outputs for automated cross-domain evaluation, reducing dependence on domain-specific parsing.
    \item \textbf{Early Learning Signal}: 
    The fixed schema offers an immediate reward in early RL training, speeding adaptation to the target format and discovery of generalizable tool-use strategies.
    \item \textbf{Explicit Termination}: Calling $\mathcal{T}_{\text{answer}}$ defines task completion, providing a domain-invariant halting signal transferable across domains (e.g., math or science tasks).
\end{itemize}

The standardized tool interface plays a pivotal role in enabling effective zero-RL training, where the model is optimized directly from a base model without prior exposure to tool-use or structured reasoning data. In this setting, the model initially lacks awareness of how to format tool invocations or when to terminate reasoning. The fixed and minimal interface design—centered around the single \texttt{answer} tool and the enforced \verb|\boxed{}| output schema—provides a strong, low-entropy learning signal early in training. 


\subsection{Verifiable Reward Mechanism via Symbolic Equivalence}

A common failure mode in reinforcement learning with language models is \emph{reward hacking}, where agents exploit superficial patterns in the reward function—such as formatting tricks or partial guesses—without producing logically correct solutions. This issue is especially severe in mathematical reasoning tasks involving tools, where incorrect intermediate reasoning can still lead to plausible-looking final answers.

To ensure that improvements reflect genuine reasoning capability, we design a verifiable reward mechanism based on symbolic equivalence.

We introduce \textbf{MathRuleGrader}, a deterministic evaluator that rigorously verifies mathematical correctness. Unlike simple string matching, MathRuleGrader integrates:
\begin{itemize}
    \item \textbf{LaTeX parsing} to extract final answers enclosed in \texttt{\textbackslash boxed\{\}};
    \item \textbf{Symbolic equivalence checking} using SymPy to verify mathematical correctness;
    \item \textbf{Code interpreter validation} to ensure consistency between intermediate computations and the final result.
\end{itemize}

This multi-layer verification ensures that only mathematically valid solutions are rewarded, regardless of superficial textual similarity.

\subsection{Specialized Prompt Template Design}\label{app:prompt-template}

We employ a specialized prompt template, \texttt{chat\_template}, to structure interactions between the model and tools. This template introduces several innovations:
\begin{itemize}
    \item \textbf{XML-Based Tool Call Format}: Tool calls are wrapped in \verb|<tool_call>...</tool_call>| tags, simplifying the model’s learning of correct syntax.
    \item \textbf{Explicit Thinking Section}: The model is encouraged to reason through problems in a dedicated \verb|<think>...</think>| block before making tool calls.
    \item \textbf{Structured Conversation Flow}: Multi-turn dialogue is organized into rounds with explicit role markers, such as \verb|[Round 0] USER:| and \verb|ASSISTANT:|.
    \item \textbf{Tool Response Integration}: Tool outputs are integrated using the \verb|TOOL:| marker for seamless information flow.
\end{itemize}

These design choices offer several advantages:
\begin{itemize}
    \item Clear structural boundaries for tool calls and reasoning steps.
    \item Encouragement of deliberate, step-wise problem-solving.
    \item Enhanced coherence in multi-turn conversations.
    \item Standardized integration of tool responses.
\end{itemize}

Empirical results show that this template significantly improves the model’s ability to format tool calls correctly and maintain coherent, multi-step interactions.

Example template snippet:

\begin{lstlisting}[basicstyle=\ttfamily\small, breaklines=true, frame=single, backgroundcolor=\color{gray!10}]
[Round 0] USER: ...
ASSISTANT: ...
<think>
reasoning process here
</think>
<tool_call>
{"name": <function-name>, "arguments": <args-dict>}
</tool_call>
TOOL: {"name": "tool_name", "content": response_content}
\end{lstlisting}



\subsection{Experimental Details}

All experiments are conducted using identical computational resources and evaluation protocols to ensure fair comparison. Models are trained on 8×A100 GPUs with mixed precision training and gradient accumulation. Evaluation is performed using consistent random seeds and identical sampling parameters across all baselines. For cross-domain evaluation, we ensure that no target domain data is used during training, maintaining strict separation between training and evaluation domains to provide unbiased assessment of generalization capabilities.

\subsection{Answer Tool Design}

We introduce a specialized \texttt{answer} tool as a critical component of our RL framework. Unlike traditional tool-calling approaches that primarily focus on interacting with external APIs, our answer tool enforces a standardized response format and serves as an explicit termination signal for the model’s reasoning process. The answer tool is defined via a structured JSON schema:
\begin{lstlisting}[basicstyle=\ttfamily\small, breaklines=true, frame=single, backgroundcolor=\color{gray!10}]
{
    "type": "function",
    "function": {
        "name": "answer",
        "description": "Respond to the user",
        "parameters": {
            "type": "object",
            "properties": {
                "answer": {"type": "string", "description": "Response content, place your final answer within \boxed{} notation."}
            },
            "required": ["answer"]
        }
    }
}
\end{lstlisting}

This design provides several advantages:
\begin{itemize}
    \item \textbf{Format Standardization}: By requiring final answers to be enclosed in \verb|\boxed{}| notation, the model outputs are consistent and easily extractable for automated evaluation.
    \item \textbf{Early Learning Signal}: The answer tool offers an immediate and clear learning signal during early RL training, accelerating the model’s adaptation to the expected output format.
    \item \textbf{Explicit Termination}: It serves as an explicit endpoint for the model’s reasoning, clearly indicating when a problem has been solved.
\end{itemize}

This mechanism significantly reduces the time required for the model to match the performance of supervised fine-tuned (SFT) baselines in terms of answer formatting and termination.

\subsection{XML-Based Prompt Template}

We employ a structured XML-based prompt template $\mathcal{P}$ to govern interactions between the model and tools, encouraging modular, domain-invariant planning and coherent multi-turn interactions. 
The template introduces three key elements:

\begin{itemize}
    \item \textbf{XML-based tool invocation}: Each tool call is enclosed within $\langle \texttt{tool\_call} \rangle \cdot \langle / \texttt{tool\_call} \rangle$, providing unambiguous syntax for learning and reducing hallucination in transfers.
    \item \textbf{Dedicated reasoning block}: Reasoning is explicitly encouraged within a $\langle \texttt{think} \rangle \cdot \langle / \texttt{think} \rangle$ section, promoting structured, step-by-step derivation before tool usage, which elicits high-level strategies (e.g., problem decomposition) applicable across domains.
    \item \textbf{Role-structured dialogue}: Interactions are segmented into rounds, denoted as
    \[
    [\text{Round } t] \quad \text{USER: } u_t, \quad \text{ASSISTANT: } a_t,
    \]
    where tool responses are incorporated via a standardized \texttt{TOOL:} marker.
\end{itemize}

This prompt template defines a clear syntax separating reasoning, tool calls, and final responses, reducing redundancy and improving token efficiency across domains. By enforcing interpretable step-by-step reasoning within defined tags, it fosters structured problem solving and narrows cross-domain generalization gaps. It also enhances coherence in multi-turn dialogues, enabling transferable reasoning strategies—e.g., applying optimization behaviors learned in math to scientific experiments. Finally, its consistent tool integration with dynamic cues supports robust evaluation of domain invariance and strengthens the agent’s adaptive tool-use across varied reasoning tasks. A complete example of the XML-based prompt template $\mathcal{P}$ used in our experiments is provided in Appendix~\ref{app:prompt-template}.

\newpage

\onecolumn

\section{Case Study}

\subsection{A case from Chemistry Problem Solving}
\begin{tcolorbox}[
colback=white,
colframe=black,
title=A case from Chemistry Problem Solving,
breakable
]
\textcolor{red}{\textit{\textbf{Problem}:}}\\
Nitrosyl chloride (\(\ce{NOCl}\)) decomposes to nitric oxide (\(\ce{NO}\)) and chlorine (\(\ce{Cl2}\)) when heated according to the reaction:

\[
\ce{2 NOCl (g) <=> 2 NO (g) + Cl2 (g)}
\]

At \(T = 600\ \mathrm{K}\), the equilibrium constant is \(K_p = 0.060\). In a vessel at \(600\ \mathrm{K}\), a mixture of all three gases has the following partial pressures:
$
P_{\ce{NOCl}} = 675\ \mathrm{torr}, \quad
P_{\ce{NO}} = 43\ \mathrm{torr}, \quad
P_{\ce{Cl2}} = 23\ \mathrm{torr}.
$
 What is the partial pressure of \(\ce{NO}\) at equilibrium?

\textcolor{red}{\textit{\textbf{Solution}:}}\\
\textcolor{blue}{\textbf{ASSISTANT:}}

This is a chemical equilibrium problem involving the decomposition of nitrosyl chloride. I need to:
\begin{enumerate}
    \item Determine the reaction direction by comparing Q (reaction quotient) with Kp
    \item Set up an ICE table to track pressure changes
    \item Solve for the equilibrium partial pressure of NO
\end{enumerate}

The balanced equation is: $2NOCl(g) \rightarrow 2NO(g) + Cl_2(g)$

Given data:
\begin{itemize}
    \item Temperature: 600 K
    \item $K_p = 0.060$
    \item Initial partial pressures: $P_{NOCl} = 675$ torr, $P_{NO} = 43$ torr, $P_{Cl_2} = 23$ torr
\end{itemize}

First, I'll calculate the reaction quotient Q to determine which direction the reaction proceeds.

\textbf{Step 1: Calculate Reaction Quotient Q}

\begin{tcolorbox}[colback=cyan!5!white, colframe=cyan!75!black, title=Python Code]
\begin{verbatim}
# Initial pressures (torr) and Kp
P_NOCl, P_NO, P_Cl2 = 675, 43, 23
Kp = 0.060

# Reaction quotient Q
Q = (P_NO**2 * P_Cl2)/(P_NOCl**2)
Q
\end{verbatim}
\end{tcolorbox}

\begin{tcolorbox}[colback=orange!5!white, colframe=orange!75!black, title=Tool Output]
\begin{verbatim}
0.09333772290809328
\end{verbatim}
\end{tcolorbox}

\textcolor{blue}{\textbf{ASSISTANT:}} Since $Q_{\text{initial}} = 0.0933 > K_p = 0.060$, the reaction will shift in the reverse direction to reach equilibrium. This means some \ce{NO} and \ce{Cl2} will combine to form more \ce{NOCl}.

\textbf{Step 2: Set up ICE Table and Solve}

Let $x$ be the extent of the reverse reaction. The changes in partial pressures will be:
\begin{itemize}
    \item $P_{NOCl}$ increases by $2x$
    \item $P_{NO}$ decreases by $2x$ 
    \item $P_{Cl_2}$ decreases by $x$
\end{itemize}

At equilibrium:
\begin{align}
P_{NOCl,eq} &= 675 + 2x \\
P_{NO,eq} &= 43 - 2x \\
P_{Cl_2,eq} &= 23 - x
\end{align}

Substituting into the equilibrium expression:
$$K_p = \frac{(P_{NO,eq})^2 \cdot P_{Cl_2,eq}}{(P_{NOCl,eq})^2} = 0.060$$

This gives us:
$$\frac{(43-2x)^2(23-x)}{(675+2x)^2} = 0.060$$

This is a cubic equation that requires numerical solution.

\begin{tcolorbox}[colback=cyan!5!white, colframe=cyan!75!black, title=Python Code]
\begin{verbatim}
import numpy as np
from scipy.optimize import fsolve

# Define the cubic equation coefficients
# Equation: 4x³ - 263.76x² + 5967x -15189.5 = 0
def func(x):
    return 4*x**3 - 263.76*x**2 + 5967*x - 15189.5

# Initial guess for x
initial_guess = 10 # Starting with a reasonable guess
solution = fsolve(func, initial_guess)

# Extract the valid solution (should be a real positive root)
x = solution[0]

x
\end{verbatim}
\end{tcolorbox}

\begin{tcolorbox}[colback=orange!5!white, colframe=orange!75!black, title=Tool Output]
\begin{verbatim}
2.901292604180679
\end{verbatim}
\end{tcolorbox}

\textcolor{blue}{\textbf{ASSISTANT:}} Now I'll calculate the equilibrium partial pressures and verify the solution.

\textbf{Step 3: Calculate Equilibrium Pressures and Verify}

\begin{tcolorbox}[colback=cyan!5!white, colframe=cyan!75!black, title=Python Code]
\begin{verbatim}
# Calculated x
x = 2.901292604180679

# Equilibrium partial pressures
P_NOCl_eq = 675 + 2*x
P_NO_eq = 43 - 2*x
P_Cl2_eq = 23 - x

# Compute Kp
Kp_calculated = (P_NO_eq**2 * P_Cl2_eq) / (P_NOCl_eq**2)
Kp_calculated
\end{verbatim}
\end{tcolorbox}

\begin{tcolorbox}[colback=orange!5!white, colframe=orange!75!black, title=Tool Output]
\begin{verbatim}
0.059999999999999984
\end{verbatim}
\end{tcolorbox}

\textcolor{blue}{\textbf{ASSISTANT:}} Perfect! The calculated $K_p \approx 0.060$ matches the given equilibrium constant, confirming our solution is correct.

\textbf{Final Answer:}

With $x = 2.901$ torr, the equilibrium partial pressure of NO is:
$$P_{NO,eq} = 43 - 2(2.901) = 37.2 \text{ torr}$$

\textbf{Verification:}
\begin{itemize}
    \item All equilibrium pressures are positive ✓
    \item $K_p$ calculation matches the given value ✓
    \item Reaction proceeds in the correct direction (reverse) ✓
\end{itemize}

Therefore, the partial pressure of NO at equilibrium is $\boxed{37.2 \text{ torr}}$.

\end{tcolorbox}

\subsection{A case from Mathematical Problem Solving}

\begin{tcolorbox}[
colback=white,
colframe=black,
title=A case from Mathematical Problem Solving,
breakable
]
\textcolor{red}{\textit{\textbf{Problem}:}}\\
Let $N$ be the greatest four-digit positive integer with the property that whenever one of its digits is changed to $1$, the resulting number is divisible by $7$. Let $Q$ and $R$ be the quotient and remainder, respectively, when $N$ is divided by $1000$. Find $Q+R$.

\textcolor{red}{\textit{\textbf{Solution}:}}\\

\textcolor{blue}{\textbf{ASSISTANT:}} I need to find the largest four-digit number $N$ such that changing any single digit to 1 results in a number divisible by 7.

\textbf{Problem Analysis:}
\begin{enumerate}
    \item Find $N$: The largest four-digit number where changing any digit to 1 gives a multiple of 7
    \item For a four-digit number $ABCD$, all of $1BCD$, $A1CD$, $AB1D$, $ABC1$ must be divisible by 7
    \item Once found, compute $Q = \lfloor N/1000 \rfloor$ and $R = N \bmod 1000$, then find $Q+R$
\end{enumerate}

\textbf{Approach:} Since we want the greatest such number, I'll search from 9999 downward until finding a valid $N$.

\begin{tcolorbox}[colback=cyan!5!white, colframe=cyan!75!black, title=Python Code]
\begin{verbatim}
def find_n():
    for n in range(9999, 999, -1):
        s = str(n).zfill(4)
        valid = True
        for i in range(4):
            # Replace ith digit with '1'
            new_s = s[:i] + '1' + s[i+1:]
            m = int(new_s)
            if m % 7 != 0:
                valid = False
                break
        if valid:
            return n
    return None

N = find_n()
Q = N // 1000
R = N % 1000
result = Q + R
print(f"N = {N}, Q = {Q}, R = {R}, Result = {result}")
\end{verbatim}
\end{tcolorbox}

\begin{tcolorbox}[colback=red!5!white, colframe=red!75!black, title=Error Output]
\begin{verbatim}
SyntaxError: unexpected character after line continuation character
\end{verbatim}
\end{tcolorbox}

\textcolor{blue}{\textbf{ASSISTANT:}} The error occurred due to JSON formatting issues with the f-string quotes. Let me fix this by using single quotes in the print statement to avoid conflicts with the JSON double quotes.

\begin{tcolorbox}[colback=cyan!5!white, colframe=cyan!75!black, title=Python Code]
\begin{verbatim}
def find_n():
    for n in range(9999, 999, -1):
        s = str(n).zfill(4)
        valid = True
        for i in range(4):
            new_s = s[:i] + '1' + s[i+1:]
            m = int(new_s)
            if m % 7 != 0:
                valid = False
                break
        if valid:
            return n
    return None

N = find_n()
Q = N // 1000
R = N % 1000
result = Q + R
print('N = {}, Q = {}, R = {}, Result = {}'.format(N, Q, R, result))
\end{verbatim}
\end{tcolorbox}

\begin{tcolorbox}[colback=orange!5!white, colframe=orange!75!black, title=Successful Output]
\begin{verbatim}
N = 5694, Q = 5, R = 694, Result = 699
\end{verbatim}
\end{tcolorbox}

\textcolor{blue}{\textbf{ASSISTANT:}} Perfect! The algorithm found $N = 5694$. Let me verify this result by checking that changing each digit to 1 produces numbers divisible by 7.

\textbf{Verification:}
\begin{itemize}
    \item Original number: $5694$
    \item Change thousands digit (5→1): $1694 = 7 \times 242$ ✓
    \item Change hundreds digit (6→1): $5194 = 7 \times 742$ ✓  
    \item Change tens digit (9→1): $5614 = 7 \times 802$ ✓
    \item Change units digit (4→1): $5691 = 7 \times 813$ ✓
\end{itemize}

All four modified numbers are indeed divisible by 7, confirming that $N = 5694$ is valid.

\textbf{Final Calculation:}
\begin{align}
N &= 5694 \\
Q &= \lfloor 5694/1000 \rfloor = 5 \\
R &= 5694 \bmod 1000 = 694 \\
Q + R &= 5 + 694 = 699
\end{align}

Therefore, $Q + R = \boxed{699}$.

\end{tcolorbox}

\subsection{A case from Financial Problem Solving}

\begin{tcolorbox}[
colback=white,
colframe=black,
title=A case from Financial Mathematics Problem Solving,
breakable
]
\textcolor{red}{\textit{\textbf{Problem}:}}\\
A solid waste disposal company borrowed money at 10\% per year interest to purchase new haulers and other equipment needed at the company-owned landfill site. If the company got the loan 2 years ago and paid it off with a single payment of \$4,600,000, what was the principal amount P of the loan?

\textcolor{red}{\textit{\textbf{Solution}:}}\\
\textcolor{blue}{\textbf{ASSISTANT:}}

This is a financial mathematics problem involving loan repayment calculations. I need to:
\begin{enumerate}
    \item Determine whether to use simple or compound interest
    \item Apply the appropriate formula to find the principal amount
    \item Verify the calculation
\end{enumerate}

Given data:
\begin{itemize}
    \item Annual interest rate: 10\%
    \item Loan duration: 2 years
    \item Single repayment amount: \$4,600,000
\end{itemize}

The problem states "10\% per year interest" without specifying compounding. In financial contexts, this could be either simple or compound interest. Let me calculate both possibilities to determine the most appropriate approach.

\textbf{Step 1: Calculate Principal Using Simple Interest}

For simple interest: $A = P(1 + rt)$, where:
\begin{itemize}
    \item $A = 4,600,000$ (total repayment)
    \item $r = 0.10$ (annual interest rate)
    \item $t = 2$ years
\end{itemize}

Solving for $P$: $P = \frac{A}{1 + rt}$

\begin{tcolorbox}[colback=cyan!5!white, colframe=cyan!75!black, title=Tool Call]
\begin{verbatim}
# Calculating principal for simple interest
A_simple = 4600000
r_simple = 0.10
t = 2
P_simple = A_simple / (1 + r_simple * t)
P_simple
\end{verbatim}
\end{tcolorbox}

\begin{tcolorbox}[colback=orange!5!white, colframe=orange!75!black, title=Tool Output]
\begin{verbatim}
3833333.3333333335
\end{verbatim}
\end{tcolorbox}

\textcolor{blue}{\textbf{ASSISTANT:}}

\textbf{Step 2: Calculate Principal Using Compound Interest}

For compound interest: $A = P(1 + r)^t$

Solving for $P$: $P = \frac{A}{(1 + r)^t}$

\begin{tcolorbox}[colback=cyan!5!white, colframe=cyan!75!black, title=Tool Call]
\begin{verbatim}
# Calculating principal for compound interest
A_compound = 4600000
r_compound = 0.10
t = 2
P_compound = A_compound / (1 + r_compound)**t
P_compound
\end{verbatim}
\end{tcolorbox}

\begin{tcolorbox}[colback=orange!5!white, colframe=orange!75!black, title=Tool Output]
\begin{verbatim}
3801652.892561983
\end{verbatim}
\end{tcolorbox}

\textcolor{blue}{\textbf{ASSISTANT:}}

\textbf{Step 3: Analysis and Final Answer}

The calculations yield different results:
\begin{itemize}
    \item Simple interest: \$3,833,333.33
    \item Compound interest: \$3,801,652.89
\end{itemize}

Since the problem doesn't specify compounding and uses the phrase "10\% per year interest," in most financial mathematics problems of this type, compound interest is the standard assumption unless explicitly stated as simple interest. This is because:
\begin{enumerate}
    \item Real-world loans typically use compound interest
    \item The 2-year duration suggests annual compounding
    \item Financial institutions standard practice
\end{enumerate}

\textbf{Verification:}
Using compound interest: $P \times (1.10)^2 = 3,801,652.89 \times 1.21 = 4,600,000$ ✓

Therefore, the principal amount of the loan is $\boxed{\$3,801,652.89}$.

\end{tcolorbox}

\newpage
\twocolumn

\section{Evaluation Benchmarks}
Our comprehensive evaluation encompasses challenging benchmarks across mathematical and general domains:

\textbf{Mathematical Reasoning Benchmarks:}
\begin{itemize}[leftmargin=2em]
\item \textbf{MATH-500}~\citep{hendrycks2021measuring}: A subset of 500 problems from the MATH dataset, featuring competition-level mathematical problems across algebra, geometry, number theory, and other mathematical domains. This benchmark requires sophisticated mathematical reasoning and problem-solving skills.

\item \textbf{AIME 24 \& AIME 25}: Problems from the American Invitational Mathematics Examination for 2024 and 2025, representing some of the most challenging high school mathematics competitions. These benchmarks test advanced mathematical reasoning capabilities and require precise computational skills.

\item \textbf{HMMT 25}: Problems from the Harvard-MIT Mathematics Tournament 2025, featuring university-level mathematical challenges that demand deep mathematical insight and creative problem-solving approaches.
\end{itemize}

\textbf{General Reasoning Benchmarks:}
\begin{itemize}[leftmargin=2em]
\item \textbf{GPQA}~\citep{rein2023gpqagraduatelevelgoogleproofqa}: Graduate-level questions across physics, chemistry, and biology that require expert-level domain knowledge. We use the highest-quality GPQA-diamond subset, which contains carefully curated questions designed to be challenging even for domain experts.

\item \textbf{TheoremQA}~\citep{chen2023theoremqatheoremdrivenquestionanswering}: A benchmark assessing models' ability to apply mathematical and scientific theorems to solve complex problems. It includes 800 high-quality questions covering 350 theorems from mathematics, physics, and other scientific domains, requiring both theoretical knowledge and practical application skills.

\item \textbf{WebInstruct}: A held-out validation split from the WebInstruct dataset~\citep{ma2025generalreasoneradvancingllmreasoning}, designed to evaluate multidisciplinary reasoning capabilities. 
\end{itemize}

\end{document}